\documentclass[letterpaper]{article} % DO NOT CHANGE THIS
\usepackage{aaai24}  % DO NOT CHANGE THIS
\usepackage{times}  % DO NOT CHANGE THIS
\usepackage{helvet}  % DO NOT CHANGE THIS
\usepackage{courier}  % DO NOT CHANGE THIS
\usepackage[hyphens]{url}  % DO NOT CHANGE THIS
\usepackage{graphicx} % DO NOT CHANGE THIS
\usepackage{natbib}  % DO NOT CHANGE THIS AND DO NOT ADD ANY OPTIONS TO IT
\usepackage{caption} % DO NOT CHANGE THIS AND DO NOT ADD ANY OPTIONS TO IT
\usepackage{algorithm}
\usepackage{algorithmic}

\usepackage{newfloat}
\usepackage{listings}
\DeclareCaptionStyle{ruled}{labelfont=normalfont,labelsep=colon,strut=off} % DO NOT CHANGE THIS
\floatstyle{ruled}
\newfloat{listing}{tb}{lst}{}
\floatname{listing}{Listing}
\usepackage{times}
\usepackage{epsfig}
\usepackage{graphicx}
\usepackage{amsmath}
\usepackage{amssymb}
\usepackage{amsthm}

\usepackage{enumitem}
\usepackage{tikz}
\usepackage{comment}

\usepackage{color}
\usepackage{colortbl}
\usepackage{multirow}
\usepackage{booktabs}
\usepackage{subcaption}
\usepackage{subfloat}
\usepackage{mathtools}
\usepackage{dsfont}
\usepackage{bm}
\usepackage[T1]{fontenc}
\usepackage{fancyhdr}

\definecolor{darkgrey}{rgb}{0.53,0.53,0.53}
\definecolor{mygrey}{rgb}{0.9,0.9,0.9}
\definecolor{purple}{RGB}{230, 227, 254}
\definecolor{lightgreen}{RGB}{238, 252, 241}
\definecolor{lightred}{RGB}{231, 187, 187}
\definecolor{darkred}{RGB}{198, 129, 129}

\definecolor{tabhighlight}{HTML}{e5e5e5}
\newtheorem{definition}{Definition}
\newtheorem{remark}{Remark}

\usepackage{microtype}
\usepackage{float}
\usepackage{placeins}
\usepackage{enumitem}
\usepackage{xstring}
\usepackage{xspace}
\usepackage{url}
\usepackage[hang,flushmargin]{footmisc}
\usepackage{array}
\usepackage{boldline}

\usepackage[utf8]{inputenc} % allow utf-8 input
\usepackage{url}            % simple URL typesetting
\usepackage{amsfonts}       % blackboard math symbols
\usepackage{nicefrac}       % compact symbols for 1/2, etc.
\usepackage{xcolor}         % colors
\usepackage{sidecap}
\usepackage[most]{tcolorbox}  
\let\oldcite\cite 
\renewcommand{\cite}[1]{\textcolor{blue}{\oldcite{#1}}}

\title{Understanding Prompt Tuning for V-L Models \\Through the Lens of Neural Collapse}

\author{
   Didi Zhu\textsuperscript{\rm 1} \quad
    Zexi Li\textsuperscript{\rm 1} \quad
   Min Zhang\textsuperscript{\rm 1} \quad
   Junkun Yuan\textsuperscript{\rm 1}, \\
    Yunfeng Shao\textsuperscript{\rm 2} \quad
    Jiashuo Liu\textsuperscript{\rm 3} \quad
    Kun Kuang\textsuperscript{\rm 1} \quad
    Yinchuan Li\textsuperscript{\rm 2} \quad
    Chao Wu\textsuperscript{\rm 1}
}
\affiliations{
    \textsuperscript{\rm 1}Zhejiang University \quad
    \textsuperscript{\rm 2}Huawei Noah's Ark Lab \quad
    \textsuperscript{\rm 3}Tsinghua University \quad

}

\usepackage{bibentry}
\begin{document}

\maketitle

\begin{abstract}
    Large-scale vision-language (V-L) models have demonstrated remarkable generalization capabilities for downstream tasks through prompt tuning. 
    However, the mechanisms behind the learned text representations are unknown, limiting further generalization gains, especially under class imbalance scenarios. 
    Recent advances in the neural collapse (NC) phenomenon of vision-only models suggest that the optimal representation structure is the simplex ETF, which paves the way to study representations in V-L models. 
    In this paper, we make the first attempt to use NC for examining the representations in V-L models via prompt tuning. It is found that NC optimality of text-to-image representations shows a positive correlation with downstream generalizability, which is more severe under class imbalance settings. 
    To improve the representations, we propose Neural-collapse-anchored Prompt Tuning (NPT), a novel method that learns prompts with text and image representations that satisfy the same simplex ETF. NPT incorporates two regularization terms: language-modality collapse and multi-modality isomorphism; and it is compatible with other prompt tuning methods. 
    Extensive experiments show that NPT can consistently help to improve existing prompt tuning techniques across 11 datasets for both balanced and imbalanced settings.
    % , achieving an absolute average gain of 2.90\% for novel classes and 2.25\% for harmonic mean.
\end{abstract}
  % \begin{figure}[tbp]

  %   \centering
  %   \subfloat{
  %   \includegraphics[width=3.1in,height=2.2in]{fig/fig1_model.pdf}
  %   % \includegraphics[width=3.1in,height=2.2in]{fig/eft_vis.pdf}
  %   }
  %   \caption{\textbf{Comparison of NPT with standard prompt learning methods.} (a) Existing methods suffer from lower generalization capabilities in the presence of class imbalance due to the lack of constraints on feature structures. (b) NPT addresses this issue by ensuring that text representations and image representations adhere to a common simplex ETF structure.
  %   }
  %   \label{fig:fig1}
  % \end{figure}

\section{Introduction}

Recent advancements in the realm of large-scale vision-language (V-L) pre-trained models such as CLIP \cite{radford2021learning}, have emerged as a promising alternative for visual representation learning.
Drawing strength from millions of image-text pairs, these models proficiently align textual and visual modalities, showcasing promising performance on various downstream tasks such as zero-shot image recognition~\cite{zhou2022learning}.

Despite these successes, emerging studies~\cite{zhou2022conditional,khattak2023maple} highlight the limitations of large V-L models in tasks where better generalization is needed, e.g., identifying novel classes. To mitigate these shortcomings without directly fine-tuning numerous parameters, many studies have devised methods for soft prompt tuning~\cite{zhou2022conditional,khattak2023maple}. 
Soft prompt tuning enhances the generalization abilities of V-L models by freezing their core parameters and exclusively refining learnable prompts using a limited dataset. Building on CoOp~\cite{zhou2022learning} that models context words in prompts as learnable vectors, recent studies have introduced various techniques to enhance the generalization by prompt tuning, including generating instance-specific prompt~\cite{zhou2022conditional} and integrating additional vision-modality prompt~\cite{khattak2023maple,yao2023visual}. 

The nature of prompt tuning is to learn text representations that are more adaptive to the image representations of the downstream tasks. Previous works attempted to hand-craft more effective prompt learning methods which seem to have better text representations, but they did not delve deeper into the mechanisms behind such representations. Therefore, in this paper, we wonder: 
% \textit{(1) What is a good text representation for text-image alignment?} \textit{(2) What is the relationship between such a good text representation and generalizability?} \textit{(3) How can we get that good representation in the embedding space?}
\textit{(1) What characterizes effective text and image representations in terms of V-L models' generalizability?} 
\textit{(2) How can we get those effective representations in the embedding space?}

An emerging discovery called neural collapse phenomenon \cite{papyan2020prevalence,yang2022we,li2022principled} in the vision-only models has shed light on the optimal structure of visual feature representations. This phenomenon demonstrates that the last-layer features within the same class collapse to their within-class mean after the model reaches zero training loss on a sufficient and balanced dataset. Additionally, the within-class means and their corresponding classifier vectors converge to the vertices of a simplex equiangular tight frame (ETF). The simplex ETF depicts the geometric structure of several vectors that have maximal pairwise angles and equal norms \cite{papyan2020prevalence}. 

\begin{figure*}[t]
    % \vspace{-0.2cm}
  \centering
  \subfloat{
    \includegraphics[width=6.8in]{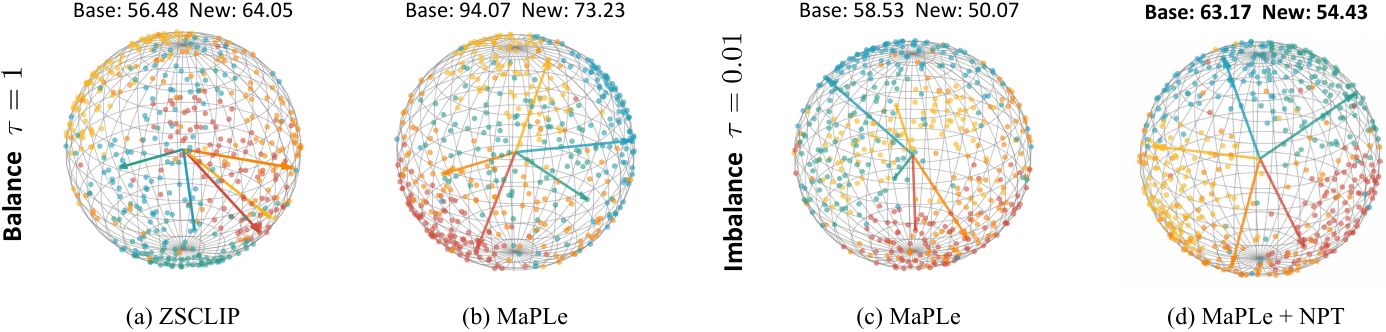}
  }
  % \vspace{-0.2cm}
  \caption{\textbf{Neural collapse degree visualization in CLIP on the EuroSAT dataset.} 
  Arrows indicate text representations, points denote image representations, and colors indicate categories. 
  % $ \tau = 1 $ for balanced data and $ \tau = 0.01 $ for high imbalance.
  (a-b) Under balanced conditions, the SOTA MaPLe method amplifies CLIP's generalizability by intensifying the neural collapse degree. (c-d) Under imbalance, the optimal structure in MaPLe is disrupted, resulting in a performance drop. Our NPT method can refine this structure, improving generalization.
  }
  \label{fig:fig1_1}
\end{figure*}

It inspires us to examine the text representation geometries in the prompt tuning of CLIP through the lens of neural collapse. 
After tuning prompts on a balanced dataset, text representations serve as classifier vectors and tend to approximate the ETF structure, as evident in Fig.~\ref{fig:fig1_1} (a-b). Intuitively, the text representations of MaPLe are closer to the simplex ETF structure (i.e., with evenly maximal angles and equal norms) and are more aligned with the image representations than ZSCLIP, and MaPLe's base-to-novel generalization is much better than ZSCLIP. We suppose that:

\begin{tcolorbox}[notitle, rounded corners, colframe=darkgrey, colback=white, boxrule=2pt, boxsep=0pt, left=0.15cm, right=0.17cm, enhanced, shadow={2.5pt}{-2.5pt}{0pt}{opacity=5,mygrey},toprule=2pt, before skip=0.65em, after skip=0.75em 
  ]
\emph{
  % {\centering  D-SGD and average-direction SAM are asymptotically equivalent.\\}
  {
    \centering 
  {
    \fontsize{8pt}{13.2pt}\selectfont 
    A better neural collapse phenomenon of text-to-image representations indicates better generalizability of prompt-tuned CLIP. 
  }
  \\
  }
  }
\end{tcolorbox}

\begin{figure*}[tbp]
  
    \centering
    \subfloat{
    \includegraphics[width=7in]{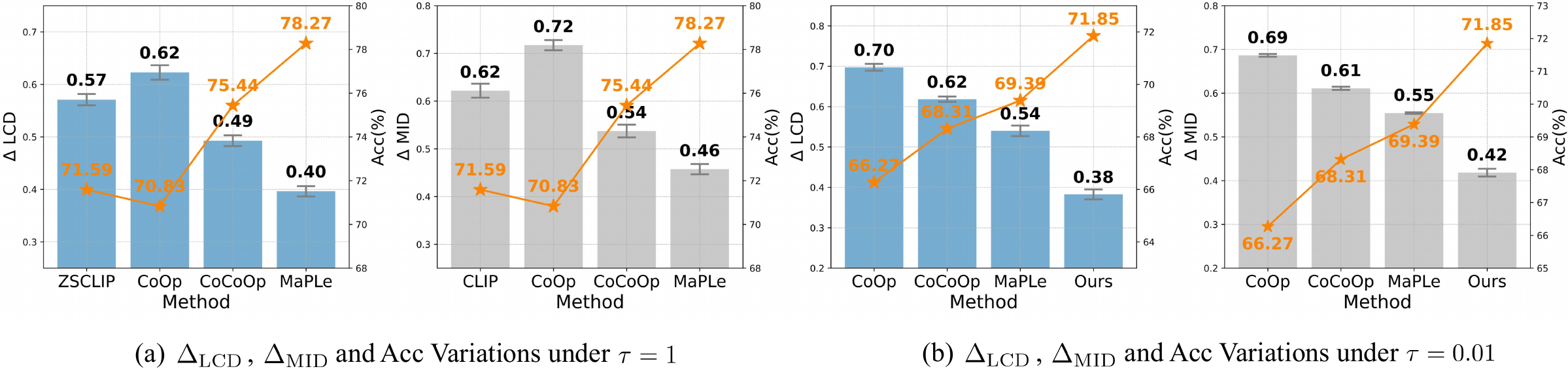}
    }
    % \vspace{-0.2cm}
    \caption{
      \textbf{Comparison of $ \Delta_\text{MID} $ and $ \Delta_\text{LCD} $ with generalization performance under the base-to-novel task.} 
      $ \tau $ denotes imbalance degree: $ \tau = 1 $ for balanced data and $ \tau = 0.01 $ for high imbalance.
      (a) In balanced settings, we compare Zero-shot CLIP with current soft prompt tuning methods, finding a direct correlation between $ \Delta_\text{MID} $ \& $ \Delta_\text{LCD}$ errors (the smaller values, the greater neural collapse) and accuracy. 
      (b) This correlation holds in class imbalance scenarios, where the degree of neural collapse and generalization performance are more impaired, even for the SOTA method MaPLe. Our approach effectively bolsters CLIP's generalization in class imbalance settings by minimizing $ \Delta_\text{LCD} $ and $ \Delta_\text{MID} $, thereby avoiding significant performance declines.
     % Under balanced training, the SOTA prompt tuning method, MaPLe, exhibits low $\Delta_\text{LCD}$ and $\Delta_\text{MID}$ values. Imbalanced data brings increased values in both metrics across 11 datasets resulting in reduced accuracy. NPT can enhance CLIP's generalization performance in class imbalance settings by effectively controlling the values of $\Delta_\text{LCD}$ and $\Delta_\text{MID}$, preventing drastic drops.
    }
    \label{fig:experiment_demo}
  \end{figure*}

It is further validated by our empirical analysis in Fig.~\ref{fig:experiment_demo} (a). It is suggested that two metrics indicative of the ETF structure (further elaborated in Sec \ref{subsec:explore}) exhibit a proportional correlation with CLIP's base-to-novel generalization performances across several prompt tuning methods. Our analysis connects the text-to-image representations of CLIP with the neural collapse phenomenon and unveils the mechanism behind the good generalizability of prompt tuning.

% 基于上述发现，我们进一步探索，发现在class imbalance的时候etf结构会被破坏，性能下降严重，为什么被破坏，class imbalance是什么，并且在现实生活中更常见，我们基于这个发现提出了一个基于nc的方法，可以通过Prompt tuning鼓励image representation和text…..

Building on the findings, we further delve into a more practical setting where the downstream dataset is class imbalanced~\cite{xie2023neural,dong2022lpt}.  
The class imbalance indicates that head classes contain most of the instances while tail classes have very few instances, which is prevalent in real-world scenarios. 
This imbalance can skew minor class representations and disturb the simplex ETF.
As demonstrated in Fig.~\ref{fig:fig1_1}~(c), minor class text representations exhibit a tendency to cluster closely, a behavior that impairs the model's capacity for generalization. This degradation of performance is further corroborated by the empirical evidence presented in Fig.~\ref{fig:experiment_demo}~(b).
% As a result, minor class text representations tend to cluster tightly, as seen in Fig.~\ref{fig:fig1_1}(c), compromising the model's generalization, further verified in Fig.~\ref{fig:experiment_demo}(b).
% when training data exhibits class imbalance, the representations of previous prompt tuning methods are more disrupted from simplex ETF, ultimately impacting the model's generalization capabilities.
To address the challenge, we propose a novel method called Neural-collapse-anchored Prompt Tuning (NPT), which aims to optimize prompts so that both text representations and image representations satisfy the same simplex ETF structure. Specifically, we introduce two regularization terms to the existing prompt tuning techniques. 
% One is a geometric de-biasing regularization term, which implicitly encourages text representations to approach the simplex ETF structure, thereby avoiding the negative impact of class imbalance. The other is a multi-modal isomorphism regularization term, which guides image representations to be closer to their corresponding text representations. 
NPT considerably enhances the generalizability of V-L models in both class-imbalanced and balanced scenarios. 

\textbf{We summarize the contributions as follows:}
    
    % \textbf{Contributions.} We summarize the primary contributions of this paper as follows: 
    \begin{itemize}[leftmargin=*]
      \item To the best of our knowledge, we are the first to extend the neural collapse phenomenon to V-L models and provide insights into the underlying representation mechanism of prompt tuning on CLIP's generalization ability.
      \item We further explore the impact of class imbalance on the generalization performance of V-L models, addressing a critical gap in existing research. To mitigate this, we introduce NPT to improve CLIP's generalization ability under both class balance and imbalance scenarios. 
      \item 
     Through theoretical analysis and extensive empirical experiments, we demonstrate the effectiveness of NPT combined with existing prompt tuning approaches, showcasing its significant improvements in generalization.
      % In base-to-novel generalization, NPT surpasses existing prompt learning techniques across 11 datasets, attaining an absolute average gain of 2.90\% for novel classes and 2.25\% for harmonic mean, in comparison to the state-of-the-art method, MaPLe \cite{khattak2023maple}. 
      % Furthermore, NPT exhibits remarkable generalization capabilities and robustness in both cross-dataset transfer and domain generalization scenarios, consistently outperforming existing methodologies.
    \end{itemize}

\textbf{Our takeaway insights of the neural collapse are below:}
    % \textbf{Contributions.} We summarize the primary contributions of this paper as follows: 
    \begin{itemize}[leftmargin=*]
      % \item A good text representation in the prompt tuning of CLIP is the representation that is close to the simple ETF of neural collapse and aligned with the image representation. If the representations approximate neural collapse better, better generalizability will be shown.
      \item  A higher degree of neural collapse in text-to-image representations is indicative of stronger generalizability. This is pivotal in explaining how soft prompt tuning enhances CLIP's generalization capability.

      \item Prompt tuning on an imbalanced dataset will disturb the text-to-image representations from neural collapse, thus, hurting the generalization. Neural-collapse-anchored approaches are needed to improve the representations.
    \end{itemize}

  \section{Related Work}
  \noindent \textbf{Vision-Language Models.}
  The fusion of language supervision with natural images has garnered significant attention within the computer vision community. These V-L models, as opposed to those solely based on image supervision, encode rich language-driven visual multi-modal representations. Recent V-L models such as CLIP \cite{radford2021learning}, ALIGN \cite{jia2021scaling}, LiT \cite{zhai2022lit}, FILIP \cite{yao2021filip}, and Florence \cite{yuan2021florence} have showcased remarkable performance across a diverse range of tasks, including few-shot and zero-shot visual recognition. These models learn joint image-language representations by self-supervised training on vast amounts of web data. 

  \noindent  \textbf{Prompt Learning.}
  % The instructions in the form of a sentence, known as a text prompt, are given to the language branch of a V-L model, allowing it to better understand the task.
  Prompt learning indicates that prompts can be learned automatically for a downstream task during the fine-tuning stage. This technique was first used in NLP~\cite{li2021prefix,lester2021power,liu2021p} followed by the adaptation in V-L~\cite{zhou2022learning,zhou2022conditional, zhu2022prompt}. 
  CoOp~\cite{zhou2022learning} fine-tunes CLIP for few-shot transfer learning by optimizing a continuous set of prompt vectors at its language branch. CoCoOp~\cite{zhou2022conditional} highlights the inferior performance of CoOp on novel classes and solves the generalization issue by explicitly conditioning prompts on image instances.
  % UPL~\cite{huang2022unsupervised} optimizes the language prompts in an unsupervised fashion.
  \cite{lu2022prompt} proposes to optimize multiple sets of prompts by learning the distribution of prompts. \cite{bahng2022visual} perform visual prompt tuning on CLIP by prompting on the vision branch. MaPLe \cite{Maaz2022Multimodal} proposes multi-modal prompts in both language and vision branches of CLIP.

  \noindent  \textbf{Neural Collapse.} 
  % Neural Collapse was initially identified in \cite{papyan2020prevalence}, revealing that during the final stage of training, a classification model trained on a balanced dataset experiences last-layer features collapsing into within-class centers. These centers and classifiers manifest as an equiangular tight frame simplex. Due to its appealing symmetry, further research endeavors to theoretically comprehend this phenomenon. It has been demonstrated that Neural Collapse is the global optimum for a simplified model under regularization \cite{zhu2021geometric, tirer2022extended}, constraint \cite{fang2021exploring, weinan2022emergence}, or without explicit constraints \cite{ji2021unconstrained}, utilizing Cross-Entropy (CE) \cite{fang2021exploring, weinan2022emergence, ji2021unconstrained, zhu2021geometric} and Mean Squared Error (MSE) loss functions \cite{tirer2022extended}. Recent studies have attempted to induce Neural Collapse in imbalanced learning \cite{yang2022we, xie2023neural,thrampoulidis2022imbalance}, incremental learning \cite{yang2023neural}, and semantic segmentation \cite{zhong2023understanding}. Nevertheless, these investigations into Neural Collapse are constrained to linear classifiers. In the present work, we uncover such solution structures present in Visual-Linguistic (V-L) models and introduce Neural Prompt Tuning (NPT) as a means to enhance the generalization capabilities of V-L models via prompt tuning.
     In \cite{papyan2020prevalence}, neural collapse was observed that a linear classification model trained on a balanced dataset experiences a phenomenon where the last-layer features collapse into within-class centers during the final stage of training. 
     % These centers and classifiers form an equiangular tight frame simplex, which displays appealing symmetry. 
    Since then, researchers have endeavored to understand this phenomenon theoretically  \cite{fang2021exploring, weinan2022emergence, ji2021unconstrained, zhu2021geometric, tirer2022extended}. 
    % It has been shown that neural collapse is the global optimum for a simplified model under regularization \cite{zhu2021geometric, tirer2022extended}, constraint \cite{fang2021exploring, weinan2022emergence}, or without explicit constraints \cite{ji2021unconstrained}. 
    Recent studies have attempted to induce neural collapse in imbalanced learning \cite{yang2022we, xie2023neural,thrampoulidis2022imbalance}, incremental learning \cite{yang2023neural}, semantic segmentation \cite{zhong2023understanding}, and federated learning \cite{li2023no,huang2023neural}. However, these investigations into neural collapse are limited to linear classifiers. In this work, we explore the presence of such a solution in V-L models and propose NPT to improve the generalizability of V-L models through prompt tuning.
  
    \section{Preliminaries}
  
  \subsection{Prompt Tuning in V-L models}
  
  \noindent \textbf{CLIP} comprises a vision-modality image encoder $G_v: \mathbb{R}^{w \times h \times 3} \rightarrow \mathbb{R}^d$ and a language-modality text encoder $G_l: \mathbb{R}^{m \times d_e} \rightarrow \mathbb{R}^{d}$. 
  The image encoder $G_v$ converts a 3-channel image $\mathbf{x}$ with shape $ w \times h$ into a $d$-dimensional image representation $\mathbf{z} \in \mathbb{R}^{d}$. 
  Meanwhile, the text encoder $G_l$ generates a $d$-dimensional text representation $\bm{g}_k \in \mathbb{R}^{d}$ from $m$-words $d_e$-dimensional text input $\mathbf{t}_k$, where $k \in \{1,...,K\}$ and $K$ is the number of classes. 
  Note that both $\mathbf{z}$ and $\bm{g}_k$ are $\ell_2$ normalized vectors.
  Two encoders are jointly trained using a contrastive loss that maximizes the cosine similarity of matched pairs and minimizes that of the unmatched pairs. 

  \noindent \textbf{Prompt engineering} adapts the pre-trained CLIP to perform zero-shot recognition without fine-tuning the model. 
  % Zero-shot transfer is enabled by formulating the classification task as an image-text matching problem. 
  Denote $\mathbb{V} = \{\texttt{\small class}_k\}_{k=1}^K$ as a class name set, the text input for CLIP is obtained by extending $\texttt{[\small class}_k\texttt{]}$ using a template like $\mathbf{t}_k = \texttt{\small ``a photo of a} \texttt{ [class}_k\texttt{]}'' $. The corresponding text representation $\bm{g}_k$ is obtained by feeding the prompt into the text encoder $G_l$, while the image representation $\mathbf{z}$ of the image $\mathbf{x}$ is extracted by the image encoder $G_v$. 
  % The prediction task is defined as the classification of an image into one of $M$ categories, which are represented by the set $y \in\{1, \ldots, M\}$. 
 % The probability of $x$ belonging to the $k$-th class is:
 %  \begin{equation}
 %  p(y=k \mid \mathbf{x})=\frac{\exp \left( \langle \mathbf{z}, \bm{g}_k\rangle / \lambda \right)}{\sum_{i=1}^K \exp \left(  \langle \mathbf{z}, \bm{g}_i\rangle / \lambda \right)},
 %  \end{equation}
 %  % \begin{equation}
 %  %   p(y=i \mid \mathbf{x})=\frac{\exp \left( \langle G_v(\mathbf{x}), G_l(\mathbf{t}_i)\rangle / \tau\right)}{\sum_{j=1}^M \exp \left(  \langle G_v(\mathbf{x}), G_l(\mathbf{t}_j)\rangle / \tau\right)},
 %  %   \end{equation}
 %  where $\langle\cdot, \cdot\rangle$ denotes the dot product operation and $\lambda$ is a temperature hyperparameter.
  
  \noindent \textbf{Soft prompt learning} replaces the hand-crafted prompts with learnable prompts. CoOp \cite{zhou2022learning} and CoCoOp \cite{zhou2022conditional} incorporate $b$ learnable tokens $\{\mathbf{u}_1, \ldots, \mathbf{u}_b\}$ for modeling text prompts, whereas 
  MaPLe \cite{Maaz2022Multimodal} extends the approach by introducing an additional set of $b$ learnable tokens $\{\mathbf{v}_1, \ldots, \mathbf{v}_b\}$ to model vision prompts.
  Denote $\bm{c}_k$ as the word embedding of the $k$-th class name, the text prompt of $k$-th class is formulated as $\mathbf{t}_k=\{\mathbf{u}_1, \ldots, \mathbf{u}_b, \bm{c}_k\}$.  For an image $\mathbf{x}$, its visual prompt is represented as $\mathbf{f} = \{\mathbf{v}_1, \ldots, \mathbf{v}_b, \mathbf{x}\}$.
  Therefore, the probability that $\mathbf{x}$ belongs to the $k$-th class can be expressed as:
  \begin{equation}
  p_\text{soft}(y=k \mid \mathbf{x})=\frac{\exp \left(\langle G_v(\mathbf{f}), G_l(\mathbf{t}_k) \rangle  / \lambda \right)}{\sum_{i=1}^K \exp \left(\langle G_v(\mathbf{f}), G_l(\mathbf{t}_i ) \rangle / \lambda \right)}.
  \label{eq:p_soft}
  \end{equation}
  Note that the prediction probability can be represented by replacing $G_v(\mathbf{f})$ with $G_v(\mathbf{x})=\mathbf{z}$ in hand-crafted prompt engineering and vision-only prompt tuning.

  % \subsection{Neural Collapse Observations}
  
  \subsection{Neural Collapse in Linear Classifier} 
% \noindent \textbf{Neural Collapse in Linear Classifier} 
% Neural collapse describes that the last-layer features will converge to their within-class means, and the within-class feature means together with the linear classifier vectors will collapse to the vertices of a simplex equiangular tight frame at the terminal training phase.
Neural collapse describes that, at the terminal training phase, the within-class means together with the linear classifier vectors will collapse to the vertices of a simplex equiangular tight frame (ETF), which is defined as follows.
  
  \begin{definition}[Simplex Equiangular Tight Frame]
  \label{def:etf}
  Consider a set of vectors $ \mathbf{m}_i  \in  \mathbb{R}^d $, with $ i = 1, \cdots, K $, and $ d \geq K-1 $. This collection of vectors forms a simplex equiangular tight frame if:
  
    \begin{equation}
    \label{eq:etf_def}
    \bm{M}=\sqrt{\frac{K}{K-1}} \mathbf{u}\left(\bm{I}_K-\frac{1}{K} \bm{1}_K \bm{1}_K^T\right),
    \end{equation}
    where $\bm{M}=\left[\mathbf{m}_1, \cdots, \mathbf{m}_K\right] \in \mathbb{R}^{d \times K}, \mathbf{u} \in \mathbb{R}^{d \times K}$ allows a rotation and satisfies $\mathbf{u}^T \mathbf{u}=\bm{I}_K, \bm{I}_K$ is the identity matrix, and $\bm{1}_K$ is an all-ones vector.
    All vectors in a simplex ETF have an equal $\ell_2$ norm and the same pair-wise angle, i.e.,
    \begin{equation}
      \label{eq:etf_prop}
    \mathbf{m}_i^T \mathbf{m}_j=\frac{K}{K-1} \delta_{i, j}-\frac{1}{K-1}, \forall i, j \in[K],
    \end{equation}
    where $\delta_{i, j}$ equals to 1 when $i=j$ and 0 otherwise. The pair-wise angle $-\frac{1}{K-1}$ is the maximal equiangular separation of $K$ vectors in $\mathbb{R}^d$ \cite{papyan2020prevalence}.
  \end{definition}
  % {Definition 1 (Simplex Equiangular Tight Frame)}
  
  Denote $\mathbf{z}_{k,i}$ as the last-layer image representation of the $i$-th sample in the $k$-th class,  the prototype $\mathbf{z}_k = \operatorname{Avg}_i\{\mathbf{z}_{k, i}\}$ is the mean value of the features in the $k$-th class, $\bm{w}_k$ is the classifier vector of the $k$-th class.
  We define three properties instructed by neural collapse in vision-only models below:

  % \begin{definition}[Feature Collapse] For any $k \in [K]$, the last-layer features in the $k$-th class collapse to their prototype, i.e, the $k$-th within-class covariance  $\Sigma_k \rightarrow \bm{0}$, where $\Sigma_k:=$ $\operatorname{Avg}_{k, i}\left\{\left(\mathbf{z}_{k, i}-\mathbf{z}_k\right)\left(\mathbf{z}_{k, i}-\mathbf{z}_k\right)^T\right\}$ and $\operatorname{Avg}\{\cdot\}$ indicates the average operation.
  % \end{definition}

  % \begin{definition}[Prototype Collapse] For any $k \in [K]$, the normalized prototypes converge to a simplex ETF, i.e., $\tilde{\mathbf{z}}_k=(\mathbf{z}_k-\mathbf{z}_G) /\|\mathbf{z}_k-\mathbf{z}_G\|$, satisfies Eq. \ref{eq:etf_prop}, where $\mathbf{z}_G$ is the global mean of the last-layer features, i.e., $\mathbf{z}_G=\operatorname{Avg}_{k, i}\{\mathbf{z}_{k, i}\}$.
  % \end{definition}

  % \begin{definition}[Classifier Collapse] For any $k \in [K]$, the normalized classifier vectors converge to the same simplex ETF as feature prototypes, i.e., $\tilde{\bm{w}}_k=\bm{w}_k /\left\|\bm{w}_k\right\| = \mathbf{z}_k$, satisfies Eq. \ref{eq:etf_prop}.
  % \end{definition}

  \noindent \textbf{Feature Collapse}: For any $k \in \{1, \dots, K\}$, the last-layer features in the $k$-th class collapse to their prototype, i.e., the $k$-th within-class covariance  $\Sigma_k \rightarrow \bm{0}$, where $\Sigma_k:=$ $\operatorname{Avg}_{k, i}\left\{\left(\mathbf{z}_{k, i}-\mathbf{z}_k\right)\left(\mathbf{z}_{k, i}-\mathbf{z}_k\right)^T\right\}$ and $\operatorname{Avg}\{\cdot\}$ indicates the average operation.

  \noindent \textbf{Prototype Collapse}: For any $k \in \{1, \dots, K\}$, the normalized prototypes converge to a simplex ETF, i.e., $\tilde{\mathbf{z}}_k=(\mathbf{z}_k-\mathbf{z}_G) /\|\mathbf{z}_k-\mathbf{z}_G\|$, satisfies Eq. \ref{eq:etf_prop}, where $\mathbf{z}_G$ is the global mean of the last-layer features, i.e., $\mathbf{z}_G=\operatorname{Avg}_{k, i}\{\mathbf{z}_{k, i}\}$.

  \noindent \textbf{Classifier Collapse}: For any $k \in \{1, \dots, K\}$, the normalized classifier vectors converge to the same simplex ETF as prototypes, i.e., $\tilde{\bm{w}}_k=\bm{w}_k /\left\|\bm{w}_k\right\| = \mathbf{z}_k$, satisfies Eq. \ref{eq:etf_prop}.
  % \noindent
  % \textbf{NC1} (Features collapse to the within-class mean): 
  % $\Sigma_W \rightarrow \bm{0}$, and $\Sigma_W:=$ $\operatorname{Avg}_{k, i}\left\{\left(\bm{h}_{k, i}-\bm{h}_k\right)\left(\bm{h}_{k, i}-\bm{h}_k\right)^T\right\}$, where $\bm{h}_k=\operatorname{Avg}_i\left\{\bm{h}_{k, i}\right\}$ is the within-class mean of the last-layer features in the $k$-th class;
  
  % \noindent
  % \textbf{NC2} (Within-class means collapse to the simplex ETF):  $\tilde{\bm{h}}_k=\left(\bm{h}_k-\bm{h}_G\right) /\left\|\bm{h}_k-\bm{h}_G\right\|, k \in[1, K]$, satisfies Eq. \ref{eq:etf_prop}, where $\bm{h}_G$ is the global mean of the last-layer features, i.e., $\bm{h}_G=\operatorname{Avg}_{k, i}\left\{\bm{h}_{k, i}\right\}$;
  
  % \noindent
  % \textbf{NC3} (Classifier vectors collapse to the same simplex ETF): $\tilde{\bm{h}}_k=\bm{w}_k /\left\|\bm{w}_k\right\|$, where $\bm{w}_k$ is the classifier vector of the $k$-th class;

  % \section{Neural Collapse in CLIP}
  % 不同数据集下 不同方法下 text representation和image representation随迭代次数增大的变化趋势
  % ZSCLIP, LPCLIP, CoOp, CoCoOp, MAPLE
  % 每个方法每个数据集对应三条曲线text representation, train image representation, test image representation
  % 比如在CoOp和CoCoOp中train/test image representation应该不会变，Maple中应该会变
  % 你想说明的是当image representation和text representation呈现出ETF同时都很明显的时候性能最好，不同方法之间对比还是很明显的

  \begin{figure}[tbp]

    \centering
    \subfloat{
    \includegraphics[width=3.4in]{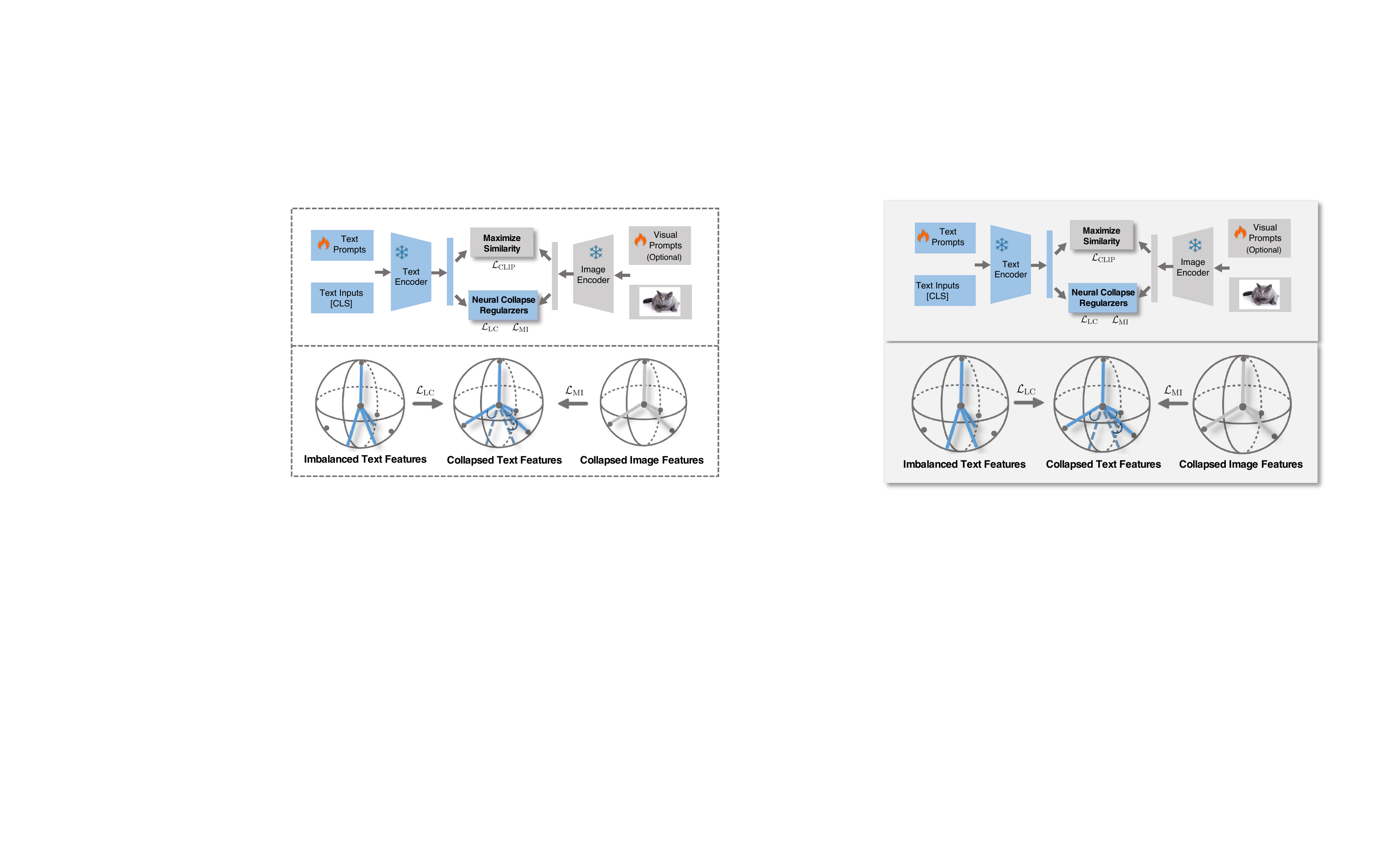}
    }
    % \vspace{-0.4cm}
    \caption{\textbf{Overview of Neural-collapse-anchored Prompt Tuning (NPT).} Our method capitalizes on the benefits of two distinct regularizers: LC Regularizer $\mathcal{L}_\text{LC}$, which controls the increase in $\Delta_\text{LCD}$ values and fosters the generation of more discriminative textual representations; and MI Regularizer $\mathcal{L}_\text{MI}$, which promotes enhanced multi-modal alignment to address challenges and attain a reduced $\Delta_\text{MID}$. 
    % As a result, this comprehensive optimization framework significantly bolsters the generalization capabilities and overall resilience of CLIP models.
    }
    \label{fig:framework}
  \end{figure}

  \section{Method}

  % \begin{figure*}[tbp]

  %   \centering
  %   \subfloat{
  %   \includegraphics[width=6.8in]{fig/framework.pdf}
  %   }
  %   \caption{\textbf{Overview of Neural-collapse-anchored Prompt Tuning (NPT) framework for prompt learning in V-L models.} Our method capitalizes on the benefits of two distinct regularizers: LC Regularizer $\mathcal{L}_\text{LC}$, which controls the increase in $\Delta_\text{LCD}$ values and fosters the generation of more discriminative textual representations; and MI Regularizer $\mathcal{L}_\text{MI}$, which promotes enhanced multimodal alignment to address challenges and attain a reduced $\Delta_\text{MID}$. 
  %   % As a result, this comprehensive optimization framework significantly bolsters the generalization capabilities and overall resilience of CLIP models.
  %   }
  %   \label{fig:framework}
  % \end{figure*}

  \subsection{Explore Neural Collapse in CLIP}
  \label{subsec:explore}

  The intriguing neural collapse (NC) phenomenon has primarily been observed and examined in vision-only models. In this paper, we investigate the corresponding structures in the vision-language model CLIP, a model that performs recognition by evaluating similarities between text and image representations. 
  In consideration of the multi-modality aspect of the V-L model, we design two evaluation criteria during training to assess the presence of neural collapse phenomena in the V-L model. The definitions are as follows:

  \begin{definition}[Language-modality Collapse Degree] Given a text encoder $G_l(\cdot)$, the language-modality collapse degree (LCD) of text representations is given by 
    $$
    \begin{aligned}
        \Delta_\text{LCD}  & = \operatorname{Avg}_{i \neq j}\left\{ \langle G_l(\mathbf{t}_i), G_l(\mathbf{t}_j)  \rangle - E_W \cdot \mu \right\}, \\
    \text { s.t. } & \left\|G_l(\mathbf{t}_i)\right\|^2 \leq E_W, \forall 1 \leq i \leq K, \\
    \end{aligned}
    $$
    
  where $\mu = -\frac{1}{K-1}$, $\mathbf{t}_i$, $\mathbf{t}_j$ are text prompts of $G_l(\cdot)$ and $\sqrt{E_W}$ is the fixed length constraint for text representations.
  \end{definition}

  \begin{definition}[Multi-modality Isomorphism Degree] Given a text encoder $G_l(\cdot)$ and an image encoder $G_v(\cdot)$, the multi-modality isomorphism degree (MID) is given by 
    % $$
    % \Delta_\text{MID} = \operatorname{Avg}_{k,i} \left\{ \langle G_v(\mathbf{f}_{k,i}), G_l(\mathbf{t}_k)  \rangle - \sqrt{E_W \cdot E_H} \right\},
    % $$
     $$
    \begin{aligned}
         \Delta_\text{MID}  & =  \operatorname{Avg}_{k,i} \left\{ \langle G_v(\mathbf{f}_{k,i}), G_l(\mathbf{t}_k)  \rangle - \sqrt{E_W \cdot E_H} \right\}, \\
    \text { s.t. } & \left\|G_l(\mathbf{t}_i)\right\|^2 \leq E_W, \forall 1 \leq i \leq K, \\
    & \left\|G_v(\mathbf{f}_{k,i})\right\|^2 \leq E_H, \forall 1 \leq k \leq K, 1 \leq i \leq n_k,
    \end{aligned}
    $$

    where $\mathbf{f}_{k,i}$ is the image input of $\mathbf{x}_i$ in the $k$-th class and $n_k$ is the number of samples belonging to the $k$-th class. $\sqrt{E_H}$ is the fixed length constraint for image features. 
  \end{definition}

\begin{remark}
       $\Delta_\text{LCD}$ measures the degree of separation among text representations (equivalent to classification vectors in vision-only models), reflecting the proximity between the language-modality structure and the simplex ETF. $\Delta_\text{MID}$ characterizes the discrepancy between the image representation and its text representation, reflecting the alignment degree of language- and vision-modality structure. Lower  $\Delta_\text{LCD}$ and $\Delta_\text{MID}$ indicate that text representations and image representations are more aligned and closer to the same simplex ETF.
\end{remark}
  
  As illustrated in Fig. \ref{fig:experiment_demo}, we analyzed the variations in $\Delta_\text{LCD}$ and $\Delta_\text{MID}$ values across different class distributions on the base-to-novel task. The reported values represent the average across 11 datasets. Furthermore, we visualize the neural collapse degree for the novel classes on the EuroSAT dataset, as depicted in Fig. \ref{fig:fig1_1}. 
  Under balanced training data conditions, the current SOTA prompt tuning method MaPLe exhibits relatively low $\Delta_\text{LCD}$ and $\Delta_\text{MID}$ values compared with ZSCLIP and other methods. This suggests a clear correlation between these values and model accuracy. However, when encountering imbalanced training data, an observable increase in both values is detected, leading to a subsequent decline in accuracy. 
  % From these findings, it is evident that class imbalance indeed hampers the generalization capability of CLIP. We try to explain the cause of this negative impact from the perspective of neural collapse. 
We can deduce that: (1) a higher degree of neural collapse, i.e., smaller values of $ \Delta_\text{LCD} $ and $ \Delta_\text{MID} $, leads to stronger generalizability, and (2) data imbalance during training disrupts the neural collapse degree, subsequently affecting generalization performance. These observations can be explicitly represented as:
  % Specifically, we conclude a correlation between imbalance degree $\tau$, $\Delta_\text{LCD}$ \& $\Delta_\text{MID}$ values, and generalization accuracy (Acc):
  \begin{equation}
    \tau \uparrow \implies   \Delta_\text{LCD} \uparrow, \Delta_\text{MID} \uparrow \implies   \text{Acc} \downarrow 
  \end{equation}
  
  Based on these observations, we propose the Neural-collapse-anchored Prompt Tuning (NPT) method, which optimizes the prompt by introducing two regularizers to implicitly address the drop in these values even under class imbalance.
As illustrated in Fig. \ref{fig:fig1_1} (d) and Fig. \ref{fig:experiment_demo} (b), our NPT enables multi-modality structures to approach the simplex ETF, thereby improving CLIP's generalization performance.
  \subsection{Neural-collapse-anchored Prompt Tuning}

  In this subsection, we propose two neural-collapse-anchored regularization terms: Language-modality Collapse (LC) Regularizer and Multi-modality Isomorphism Degree (MI) Regularizer, which are guided by $\Delta_\text{LCD}$ and $\Delta_\text{MID}$ respectively.

  \noindent \textbf{Language-modality Collapse Regularization} is introduced to address the issue in language-modality where certain text representations are situated exceptionally close while others remain distantly apart. Achieving an equitable and maximal separation among text representations is the key to the generalizability of CLIP. Hence, the LC regularizer encourages the similarity between any two text representations from different classes to approach $\mu$ (specified in Definition~\ref{def:etf}):
  % \begin{equation}
  %   \mathcal{L}_{\text{LC}} = \sum_{i,j=1, i \neq j}^{K} \left(\langle G_l(\mathbf{t}_i), G_l(\mathbf{t}_j  \rangle  - E_W \cdot \mu)\right)^2. 
  % \end{equation}
  \begin{equation}
    \mathcal{L}_{\text{LC}} =  \sum_{i=1}^{K} \sum_{j=1, i \neq j}^{K} \left(\langle G_l(\mathbf{t}_i), G_l(\mathbf{t}_j  \rangle  - E_W \cdot \mu)\right)^2. 
  \end{equation}
  where $\|G_l(\mathbf{t}_i)\|^2 \leq E_W, i = 1, \dots, K $. 
  $E_W$ is the class-wise constraints for text representation $G_l(\mathbf{t}_i)$.
  % Here, the scalar value $\mu$ is set to $-\frac{1}{K-1}$, 
  % where $K$ is the number of classes. 
  % By minimizing the LC loss term, we seek to ensure the language modality to approach the simplex ETF, thereby promoting better generalization capability across various tasks and reducing the tendency toward language modality bias.

  \noindent \textbf{Multi-modality Isomorphism Regularization} aims to tackle the issue of multi-modality isomorphism degree (MID) by optimizing the alignment between language and visual modalities as follows:
  \begin{equation}
    \mathcal{L}_{\text{MI}}=\sum_{k=1}^K \sum_{i=1}^{n_k} \left( \langle G_v(\mathbf{f}_{k,i}), G_l(\mathbf{t}_k)  \rangle - \sqrt{E_W \cdot E_H} \right)^2,
  \end{equation}
   where $\|G_v(\mathbf{f}_{k,i})\|^2 \leq E_H, i = 1, \dots, K $. 
  $E_H$ is the instance-wise constraints for image representation $G_v(\mathbf{f}_{k,i})$.
% where $B$ is the batch size and $k$ is the class index of $\mathbf{x}_i$.
  % where $\mu = -\frac{1}{K-1}$ and $p_i=\left\{v_1, \ldots, v_k, c_i\right\}$ with $k$ learnable context vectors $\left\{v_1, \ldots, v_k\right\}$.
By minimizing the MI loss term, our goal is to foster strong alignment between language and visual representations, which in turn improves model performance when handling diverse and imbalanced data distributions.
\begin{table*}[t]
  \caption{\textbf{Harmonic mean values (\%) on 11 datasets with different imbalance ratios on the base-to-novel task.}}
   % \vspace{-0.4cm}
  \label{tab:b2n}
  \centering
  \scriptsize
  \resizebox{1\linewidth}{!}{
  \renewcommand{\arraystretch}{0.7}
  \begin{tabular}{l|cccccccccccc}

      % \multirow{2}{*}{Methods} & \multicolumn{5}{c||}{PACS dataset (\%)} & \multicolumn{5}{c}{Office-Home dataset (\%)} \\
      % \cline{2-11}
      % & Art & Cartoon & Photo & Sketch & Average & Art & Clipart & Product & Real-World & Average \\

      \bottomrule

      \toprule
      \multirow{3}{*}{Method} &  \multicolumn{12}{c}{Balance with $\tau=1$}\\
      % \cline{2-14}
      % \midrule
      & 
      \multicolumn{1}{c}{IN.} &
      \multicolumn{1}{c}{Cal.} &
      \multicolumn{1}{c}{OP.} &
      \multicolumn{1}{c}{SC.} &
      \multicolumn{1}{c}{Flw.} &
      \multicolumn{1}{c}{Food.} &
      \multicolumn{1}{c}{FA.} &
      \multicolumn{1}{c}{SUN.} &
      \multicolumn{1}{c}{DTD} &
      \multicolumn{1}{c}{ES.} &
      \multicolumn{1}{c}{UCF.} &
      \multicolumn{1}{c}{Avg}  

      \\  
      
          % &Base        &Novel        &Base        &Novel         &Base        &Novel         &Base        &Novel        &Base        &Novel        &Base        &Novel        \\ 
      % Methods & Art & Cartoon & Photo & Sketch & Average \\
      \midrule
      {CLIP}   &70.22 &95.40 & 94.12 & 68.65  & 74.83  & 90.66 & 31.09 & 72.23 & 56.37  & 60.03 & 73.85 & 71.58           \\
      % {CoOp}   & 71.92 & 93.73 &94.47 &68.13 & 74.06 & 85.19 & 28.75 & 72.51 & 54.24& 68.69 & 67.46 & 70.83            \\ 
  {CoCoOp} &73.10 & 95.84 & 96.43 & 72.01 & 81.71 & 90.99 & 27.74 & 78.27 & 64.85 & 71.21 & 77.64 & 75.44           \\
  % {ProGrad}   & & & & & & & & & & & &            \\
  % {KgCoOp}   & & & & & & & & & & & &            \\
  {MaPLe}  & 73.47 & 96.02 & 96.58 & \textbf{73.47} & 82.56 & 91.38 & \textbf{36.50} & 79.75 & 68.16 & 82.35 & 80.77 &   78.27                       \\

  % \midrule
  % \rowcolor{tabhighlight}
  % {CoCoOp+NPT}  & & & & & & & & & & & &            \\ 
  \rowcolor{tabhighlight}
  {CoCoOp + NPT}  & 73.78 & \textbf{97.05} & \textbf{96.88} &72.50 & \textbf{83.01} & 91.20 &35.03 &78.67 & 66.88& 74.52& 78.82 & 77.12            \\ 
  \rowcolor{tabhighlight}
  {MaPLe + NPT}  & \textbf{74.02} & 96.46 & 96.54 & 72.71 & 82.94 & \textbf{91.75} & 35.54 & \textbf{79.88} & \textbf{69.20} & \textbf{82.85} & \textbf{81.42} & \textbf{78.48}    \\
  % {$\Delta$}  & & 96.46& 96.54& 72.71 &82.94 &90.75 &34.54 &  & 67.20 & 81.05& 81.42&      \\
  % \rowcolor{tabhighlight}
  % $\Delta$&  &  \textcolor{blue}{\textbf{+0.70}}&  \textcolor{blue}{\textbf{+0.27}} &  \textcolor{red}{\textbf{-0.18}} &  \textcolor{blue}{\textbf{+0.61}} &  \textcolor{red}{\textbf{-0.55}} &  \textcolor{red}{\textbf{-0.73}} &  \textcolor{blue}{} &  \textcolor{blue}{\textbf{+2.11}}  &  \textcolor{red}{\textbf{-1.30}} &  \textcolor{blue}{\textbf{+0.84}} &  \textcolor{blue}{\textbf{+0.21}}\\
  \bottomrule

  \toprule

  \multirow{3}{*}{Method} &  \multicolumn{12}{c}{Imbalance Ratio $\tau = 0.05$}\\
  % \cline{2-14}
  & 
  \multicolumn{1}{c}{IN.} &
  \multicolumn{1}{c}{Cal.} &
  \multicolumn{1}{c}{OP.} &
  \multicolumn{1}{c}{SC.} &
  \multicolumn{1}{c}{Flw.} &
  \multicolumn{1}{c}{Food.} &
  \multicolumn{1}{c}{FA.} &
  \multicolumn{1}{c}{SUN.} &
  \multicolumn{1}{c}{DTD} &
  \multicolumn{1}{c}{ES.} &
  \multicolumn{1}{c}{UCF.} &
  \multicolumn{1}{c}{Avg}  

  \\

  \midrule
  % {CoOp} & 68.34  & 93.02 & 93.33 & 68.56  & 72.66  & 87.38  & 18.49 & 76.43 & 52.91 & 63.57&  69.78 & 69.50 \\
  % {CoOp} & & & & & & & & & & & &                \\
{CoCoOp}  & 70.34 & 95.29 & 95.11 & 69.48 & 75.03 & 89.61 & 21.78 & 74.67 & 57.94 & 62.33 & 72.83 & 71.31 \\
% {ProGrad}   & & & & & & & & & & & &            \\
% {KgCoOp}   & & & & & & & & & & & &            \\
% {MaPLe}  & 75.41& 92.72& 89.22& 67.12& 71.84& 85.61&25.90 & 76.77 &54.85 & 46.00& 73.30& 68.98               \\
{MaPLe}  &  70.93& 95.38 & 94.22 & 70.25 & 76.59 & 88.71 & 28.62 & 75.98 & 55.15 & 51.10 & 75.43&    71.12             \\
% \midrule
\rowcolor{tabhighlight}
{CoCoOp + NPT}  & 71.21 &  \textbf{95.95} & 95.97 & 70.69 & 75.72 & 90.13 & 23.28 & 75.67 & 59.69 & 63.78 & 74.33 & 72.40 \\
% {CoCoOp+NPT} & &95.80 & 97.50& 76.87 &75.50 & 92.37& 30.63& &57.57 & 76.40& 77.70&  75.59           \\ 
\rowcolor{tabhighlight}
% {MaPLe + NPT} & 76.79&\textbf{93.07} &\textbf{90.62} &\textbf{67.99} &\textbf{73.37} &\textbf{87.24} &\textbf{26.84} & 77.94 &\textbf{55.24} &\textbf{50.74} &\textbf{74.58} &  70.40             \\
{MaPLe + NPT} &   \textbf{72.28} & 95.82 &  \textbf{96.31} &  \textbf{71.07} &  \textbf{78.76} &  \textbf{90.47} &  \textbf{30.24} &  \textbf{76.94} &  \textbf{57.35} &  \textbf{58.14} & \textbf{76.65}  &    \textbf{73.09}          \\
% \rowcolor{tabhighlight}
%     $\Delta$ &  &  \textcolor{blue}{\textbf{+0.35}}&  \textcolor{blue}{\textbf{+1.40}} &  \textcolor{blue}{\textbf{+0.87}} &  \textcolor{blue}{\textbf{+1.53}} &  \textcolor{blue}{\textbf{+1.63}} &  \textcolor{blue}{\textbf{+0.96}} &  \textcolor{blue}{} &  \textcolor{blue}{\textbf{+0.39}}  &  \textcolor{blue}{\textbf{+4.74}} &  \textcolor{blue}{\textbf{+1.28}} &  \textcolor{blue}{\textbf{+1.45}}\\
  \bottomrule

  \toprule
  \multirow{3}{*}{Method} &  \multicolumn{12}{c}{Imbalance Ratio $\tau = 0.01$}\\
  % \midrule
  % \cline{2-14}
  & 
  \multicolumn{1}{c}{IN.} &
  \multicolumn{1}{c}{Cal.} &
  \multicolumn{1}{c}{OP.} &
  \multicolumn{1}{c}{SC.} &
  \multicolumn{1}{c}{Flw.} &
  \multicolumn{1}{c}{Food.} &
  \multicolumn{1}{c}{FA.} &
  \multicolumn{1}{c}{SUN.} &
  \multicolumn{1}{c}{DTD} &
  \multicolumn{1}{c}{ES.} &
  \multicolumn{1}{c}{UCF.} &
  \multicolumn{1}{c}{Avg}  

  \\  
  
      % &Base        &Novel        &Base        &Novel         &Base        &Novel         &Base        &Novel        &Base        &Novel        &Base        &Novel        \\ 
  % Methods & Art & Cartoon & Photo & Sketch & Average \\
  \midrule
    % CoOp&65.70&93.89&91.67&66.93&70.27&89.57&10.93&71.06&48.63&50.07&70.23&66.27 \\
  % {CoOp}   & & & & & & & & & & & &            \\
  % {CoCoOp} & 73.34 & 82.42& 96.45&53.62 & 57.44&90.81 &33.36 & 73.34  & 63.66& 69.87&75.91  & 70.02          \\
  % {ProGrad}   & & & & & & & & & & & &            \\
  % {KgCoOp}   & & & & & & & & & & & &            \\
  % {MaPLe} & 73.87 & 82.83& 94.21& 53.97& 57.56&89.74 &21.89 & 75.67 &47.37 &53.97 & 73.85& 65.90             \\
  {CoCoOp} & 68.88  & 94.54& 94.71&68.45 & 71.24&88.31 &11.48 & 73.87  & 52.39& 57.45& 70.11  & 68.31          \\
  {MaPLe} & 70.10 & 95.53 & 94.21& 69.29 & 72.29 &89.74 &21.89 & 75.00 &47.37 &53.97 & 73.85& 69.38             \\
  
  % \midrule
  \rowcolor{tabhighlight}
  {CoCoOp + NPT}  &70.23  & 95.16 &  \textbf{95.66} & 70.17 & 73.50 & 90.08 & 20.34 &  74.42 & 54.97 & 62.95 & 71.01   & 70.77       \\ 
  % {CoCoOp+NPT}  & & & & & & & & & & & &          \\ 
  \rowcolor{tabhighlight}
  {MaPLe + NPT}  &  \textbf{70.89} &  \textbf{95.64} & 94.56 &  \textbf{70.58} &  \textbf{74.46} &  \textbf{90.20} &  \textbf{27.72} &  \textbf{75.89}  &  \textbf{56.98} &  \textbf{58.48} &  \textbf{74.95} &  \textbf{71.85}             \\
  % {MaPLe + NPT}  & 75.89 &\textbf{82.90} &\textbf{94.56} &\textbf{54.58} &\textbf{58.78} &\textbf{90.20} &\textbf{26.72}& 77.38 &\textbf{55.98} &\textbf{58.48} &\textbf{73.46} & 68.08              \\
  % \rowcolor{tabhighlight}
  % $\Delta$&  &  \textcolor{blue}{\textbf{+0.07}}&  \textcolor{blue}{\textbf{+0.35}} &  \textcolor{blue}{\textbf{+0.61}} &  \textcolor{blue}{\textbf{+1.22}} &  \textcolor{blue}{\textbf{+0.46}} &  \textcolor{blue}{\textbf{+4.83}} &  \textcolor{blue}{} &  \textcolor{blue}{\textbf{+8.61}}  &  \textcolor{blue}{\textbf{+4.51}} &  \textcolor{red}{\textbf{-0.39}} &  \textcolor{blue}{\textbf{+2.25}}\\

  % \bottomrule
  \bottomrule

  \toprule

  \end{tabular}}
  \end{table*}

  \begin{figure*}[htbp]

    \centering
    \subfloat{
    \includegraphics[width=7.1in,height=1.3in]{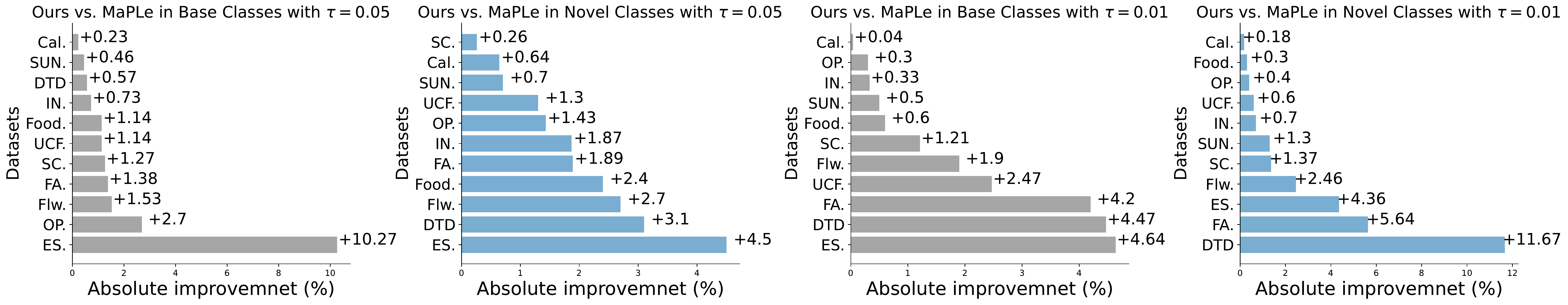}
    }
    % \vspace{-0.4cm}
    \caption{\textbf{Absolute improvement of NPT over MaPLe in the base-to-novel generalization task.} Compared with MaPLe, our method achieves improvement for both base and new classes on all datasets with $\tau = 0.05$ and $\tau = 0.01$.
    }
    \label{fig:b2n}
  \end{figure*}

\noindent \textbf{Overall Framework.}
  % Based on Eq. \ref{eq:p_soft}, the contrastive loss in soft prompt learning can be expressed:
  % \begin{equation}
  %   \label{eq:ce}
  %   \mathcal{L}_{\text{CE}}=-\frac{1}{N} \sum_{\boldsymbol{x} \in \mathcal{X}} \sum_{k=1}^K y_{\mathbf{x}, k} p_{\text{soft}}(y=k \mid \mathbf{x}),
  % \end{equation}
  By incorporating our proposed regularization terms into the contrastive loss function within the current prompt learning techniques, we can optimize the prompts effectively using the overall loss:
  \begin{equation}
    \mathcal{L} = \mathcal{L}_{\text{CLIP}} + w_1 \mathcal{L}_{\text{LC}} + w_2 \mathcal{L}_{\text{MI}}
    \label{eq:overal_loss}
  \end{equation}
  where $ \mathcal{L}_{\text{CLIP}}=-\frac{1}{N} \sum_{\boldsymbol{x} \in \mathcal{X}} \sum_{k=1}^K y_{\mathbf{x}, k} p_{\text{soft}}(y=k \mid \mathbf{x})$, 
  $w_1$ and $w_2$ are hyperparameters governing the contribution of each component.

  As illustrated in Fig. \ref{fig:framework}, the proposed approach harnesses the strengths of both regularizers: LC Regularizer for minimizing $\Delta_\text{LCD}$ value and developing more discriminative text representations, and MI Regularizer for facilitating better multi-modality alignment to achieve a lower $\Delta_\text{MID}$.
  % $$
  % \mathcal{L}\left(z, p_i\right)=\frac{1}{2}\left(G(p_i)^T z-1\right)^2
  % $$

  \subsection{Theoretical Analysis}

  \textbf{Gradient w.r.t. text prompt under class imbalance.} We first analyze the gradient of $\mathcal{L}_{\mathrm{CLIP}}$ w.r.t text representations:

%   $$
% \begin{aligned}
% & \frac{\partial \mathcal{L}_{\mathrm{CLIP}}}{\partial \mathbf{t}_k}=\frac{\partial \mathcal{L}_{\mathrm{CLIP}}}{\partial G_l\left(\mathbf{t}_k\right)} \times \frac{\partial G_l\left(\mathbf{t}_k\right)}{\partial \mathbf{t}_k}, \\
% \end{aligned}
% $$

$$
\frac{\partial \mathcal{L}_{\mathrm{CLIP}}}{\partial G_l(\mathbf{t}_k)} =\underbrace{\sum_{i=1}^{n_k}\left(p_k\left(\mathbf{z}_{k, i}\right)-1\right) \mathbf{z}_{k, i} }_{\text {intra-class cohesion}}+\underbrace{\sum_{k^{\prime} \neq k}^K \sum_{j=1}^{n_{k^{\prime}}} p_k\left(\mathbf{z}_{ k^{\prime},j}\right) \mathbf{z}_{k^{\prime},j}}_{\text {inter-class repulsion}},
$$

  where 
  $\mathbf{z}_{k,i} = G_v(\mathbf{f}_{k,i})$ , 
  $p_k(\mathbf{z}_{k,i}) = p_\text{soft}(y=k \mid \mathbf{x}_{k,i})$ is the predicted probability that $\mathbf{z}_{k,i}$ belongs to the $k$-th class.

\begin{remark}
    The gradient equation is composed of two terms. The "intra-class cohesion" term comprises $n_k$ elements, directs $ G_l(\mathbf{t}_k) $ towards its corresponding image representations $ \mathbf{z}_{k,i}$. In contrast, the "inter-class repulsion" term, with $ N-n_k $ elements, pushes $ G_l(\mathbf{t}_k) $ against features of other classes. 
    Consequently, gradients of certain minor classes are overwhelmingly influenced by the repulsion term due to the small $ n_k $ and large $ N-n_k $.
\end{remark}
   % $$
   % p_k(\mathbf{z}_{k,i}) =   \frac{\exp \left(\langle G_v(\mathbf{f}), G_l(\mathbf{t}_k) \rangle  / \lambda \right)}{\sum_{i=1}^K \exp \left(\langle G_v(\mathbf{f}), G_l(\mathbf{t}_i ) \rangle / \lambda \right)}.
   % $$ 
  To tackle this issue, we introduce $\mathcal{L}_{\text{LC}}$. The corresponding gradient equation w.r.t. $\mathbf{t}_i$ is defined as:
   $$
   \frac{\partial \mathcal{L}_{\text{LC}}}{\partial \mathbf{t}_i} = \sum_{j=1, i \neq j}^K 2\left(S_{i j}-E_W \cdot \mu\right)\left\langle\frac{\partial G_l\left(\mathbf{t}_i\right)}{\partial \mathbf{t}_i}, G_l\left(\mathbf{t}_j\right)\right\rangle,
   $$
   % $$
   %  \frac{\partial \mathcal{L}_{\mathrm{MI}}}{\partial \boldsymbol{t}_k}=\sum_{i=1}^{n_k} 2 \left(S'_{i k} - \sqrt{E_W E_H} \right)\left\langle G_v\left(\boldsymbol{f}_{k, i}\right), \frac{\partial G_l\left(\boldsymbol{t}_k\right)}{\partial \boldsymbol{t}_k}\right\rangle,
   %  $$

  where $S_{i j}=\left\langle G_l(\mathbf{t}_i), G_l(\mathbf{t}_j)\right\rangle$, $S^{\prime}_{i k}=\left\langle G_v(\mathbf{f}_{k,i}), G_l(\mathbf{t}_k)\right\rangle$.
$\mathcal{L}_{\text{LC}}$ can explicitly adjust the prompt $ \mathbf{t}_{i}$ to ensure text representations are maximally angularly separated, i.e.,  $S_{i j}-E_W \cdot \mu$. This regularizer effectively mitigates the adverse effects of imbalanced gradient updates in $\mathcal{L}_{\text{CLIP}}$.
% By fine-tuning $ \mathbf{t}_{i} $, these regularizers strategically orient text and image representations towards forming the same ETF.

 \noindent\textbf{Gradient w.r.t. image prompt.} The gradient of $\mathcal{L}_{\mathrm{CLIP}}$ with respect to the image representation $\mathbf{z}_i$ is:
 $$
 \frac{\partial \mathcal{L}_{\mathrm{CLIP}}}{\partial G_v(\mathbf{f}_i)}= (p_k(\mathbf{z}_{k,i})-1)G_l(\mathbf{t}_k)+ \sum_{k^\prime \neq k}^Kp_{k^\prime}(\mathbf{z}_{k,i})G_l(\mathbf{t}_{k^{\prime}}).
 $$
where $\mathbf{z}_{k,i} = G_v(\mathbf{f}_{k,i})$, $k$ is the class label of $\mathbf{z}_{k,i} $.
 \begin{remark}
This gradient formula is similar to $\frac{\partial \mathcal{L}_{\mathrm{CLIP}}}{\partial G_l(\mathbf{t}_k)}$. It aims to bring image representations closer to text representations of the same class and farther away from those of other classes. However, text representations can be influenced by imbalance, causing the representations of minor classes to be close together. Optimizing image representations towards biased text representations can compromise performance.
 \end{remark}
Similarly, we analyze the gradient of $\mathcal{L}_{\mathrm{MI}}$ w.r.t. $\mathbf{f}_{k, i}$:
 $$
\frac{\partial \mathcal{L}_{\mathrm{MI}}}{\partial \mathbf{f}_{k, i}}=2 \left(S'_{i j} - \sqrt{E_W E_H} \right)\left\langle\frac{\partial G_v\left(\mathbf{f}_{k, i}\right)}{\partial \mathbf{f}_{k, i}}, G_l\left(\mathbf{t}_k\right)\right\rangle.
$$
$\mathcal{L}_{\text{MI}}$ ensures that image representations closely align with text representations of the same class, uninfluenced by imbalances, as represented by $S'_{i k} - \sqrt{E_W E_H}$. This bolsters the discriminative power, especially for minor classes.
% $\mathcal{L}_{\mathrm{MI}}$ aims to align image representations with text representations by encouraging them to satisfy the same simplex ETF.

  \section{Experiment}
  Our approach is evaluated on (1) base-to-novel class identification, (2) cross-dataset transfer, and (3) domain generalization, under three class imbalance degrees. 
  All trained data are created by downsampling each class's samples to obey an exponential decay with three imbalance ratios $\tau = \{1,0.05,0.01\}$. Here $\tau =\min \left\{n_k\right\} / \max \left\{n_k\right\}$ with $n_k$ being the number of samples in the $k$-th class and $\max \left\{n_k\right\}=16$. 
  % These three problems are evaluated on both balance data and imbalance data.
  % All models used are based on the open-source CLIP. 
  % Before discussing the results, we provide the  details of the experimental setup below.

  \noindent \textbf{Datasets.} 
  For the first two settings, i.e., base-to-novel generalization and cross-dataset transfer, we use the 11 image recognition datasets as in \cite{zhou2022learning,zhou2022conditional}.  
  For domain generalization experiments, we use ImageNet as the source dataset and four other variants of ImageNet that contain different types of domain shifts as the target dataset, following \cite{zhou2022conditional}. See more details in the Appendix.

\noindent \textbf{Implementation Details.} 
Since our method is orthogonal to the other prompt tuning methods, we combine NPT with the existing CoCoOp and MaPLe, to demonstrate its flexibility and effectiveness. 
% We label these combinations as "MaPLe+NPT" and "CoCoOp+NPT", respectively.
% We use a few-shot training strategy in all experiments at 16 shots which are randomly sampled for each class. 
We apply prompt tuning on ViT-B/16 CLIP where $d=512$. 
% For MaPLe, we set prompt depth $J$ to 9 and the language and vision prompt lengths to 2. 
All models are trained for 5 epochs with a batch size of 4 and a learning rate of 0.0035 via SGD optimizer.
% We report base and novel class accuracies and their harmonic mean (HM) averaged over 3 runs.
Both $E_W$ and $E_H$ are set at $1$, while $w_1$ and $w_2$ are assigned as $0.3$ and $0.8$, respectively.
We follow other default hyperparameters used in MaPLe \cite{Maaz2022Multimodal}.
% We initialize the language prompts of the first layer $P_{0}$ with the pre-trained CLIP word embeddings of the template `\txt{a photo of a $<$category$>$}', while for the subsequent layers they are randomly initialized from a normal distribution. 
% For training MaPLe on all 1000 classes of ImageNet as a source model, prompt depth $J$ is set to 3 and the model trained for 2 epochs with learning rate of 0.0026.
% Hyper-parameters for deep language prompting, deep vision prompting, and independent V-L prompting are detailed in Appendix \ref{appendix:iml_details}. 
% The hyper-parameters are fixed across all datasets.

  \subsection{Base-to-Novel Generalization}
  Following the previous works~\cite{zhou2022conditional, khattak2023maple}, for each dataset, we split the classes equally into two groups: the base classes and the novel classes. The learnable modules are trained exclusively on the imbalanced base classes,  while evaluation is carried out separately on both the base and novel classes to testify to generalization ability.  Table \ref{tab:b2n} presents the harmonic mean values for 11 datasets on the base-to-novel tasks. Detailed base and novel accuracy figures can be found in the Appendix.
  When trained on imbalanced base classes, both CoCoOp and MaPLe see performance dips of 4.13\% and 7.15\% respectively at $ \tau =0.05 $, and 7.13\% and 8.89\% at $ \tau =0.01 $, underscoring the catastrophic impact of class imbalance on generalization.
% Both CoCoOp and MaPLe experience a decline in performance when trained on imbalanced base classes. Specifically, their performances drop by 4.13\% and 7.15\% respectively at $ \tau =0.05 $, and by 7.13\% and 8.89\% at $ \tau =0.01 $.
 % This highlights the catastrophic impact of class imbalance on generalization. 
 % In contrast, our method alleviates this impact by ensuring that the text representations and image representations satisfy the same simplex ETF structure. 
 Our method exhibits improvements of 1.97\% when $\tau =0.05$, and improvements of 2.47\% when $\tau =0.01$ over MaPLe.
%  Specifically, at an imbalance ratio of 0.05, our method exhibits an improvement of 1.09\% and 1.97\% over CoCoOp and MaPLe, respectively, while at an imbalance ratio of 0.01, it exhibits an improvement of 2.46\% and 2.47\%, respectively. 

\begin{table*}[!t]
  \scriptsize
  \caption{ \textbf{Comparison of NPT with existing approaches on cross-dataset task.}}
  % \vspace{-0.4cm}
  \label{tab:xd}
  \centering
  \resizebox{1\linewidth}{!}
  {
  \renewcommand{\arraystretch}{0.7}
  \begin{tabular}{l|cccccccccccc}
  \bottomrule
  
%   \toprule
%   & \multicolumn{12}{c}{Imbalance Ratio $\tau$ = 1} \\
%   % & \textbf{Source} & \multicolumn{11}{c}{\textbf{Target}} \\ 
%   % \cmidrule(lr){2-2} \cmidrule(lr){3-13}
%   & 
%   \multicolumn{1}{c}{IN. $\rightarrow$} &
%     \multicolumn{1}{c}{Cal.} &
%     \multicolumn{1}{c}{OP.} &
%     \multicolumn{1}{c}{SC.} &
%     \multicolumn{1}{c}{Flw.} &
%     \multicolumn{1}{c}{Food.} &
%     \multicolumn{1}{c}{FA.} &
%     \multicolumn{1}{c}{SUN.} &
%     \multicolumn{1}{c}{DTD} &
%     \multicolumn{1}{c}{ES.} &
%     \multicolumn{1}{c}{UCF.} &
%     \multicolumn{1}{c}{Avg}  
%   \\
%   \midrule
%   % CLIP &68.63 & 89.36 & 88.99 & 65.67 & 70.49 & 89.23 & 27.12 & 65.29 & 46.02 & 54.17 & 69.83 &  66.62 \\
%   CoOp & 71.51 & 93.70 & 89.14 & 64.51 & 68.71 & 85.30 & 18.47 & 64.15 & 41.92 & 46.39 & 66.55 & 63.88 \\
%   CoCoOp & 71.02 & 94.43 & 90.14 & 65.32 & 71.88 & 86.06 & 22.94 & 67.36 & 45.73 & 45.37 & 68.21 & 65.74 \\
%   MaPLe & 70.72 & 93.53 &  90.49 & 65.57 & 72.23 & 86.20 & 24.74 & 67.01 & 46.49 & 48.06 & 68.69 & 66.30 \\
%   \midrule
%   \rowcolor{tabhighlight}
% {CoCoOp + NPT}  &  \textbf{71.22} &  \textbf{94.58} &  \textbf{91.22} &  \textbf{66.91} & 71.82 &  \textbf{86.45} &  24.34& 67.88& 45.89&46.71 & \textbf{69.13} & 66.49           \\ 
% \rowcolor{tabhighlight} 
% {MaPLe + NPT} & 70.70 & 94.50 & 90.40 & 66.73 & \textbf{72.33}   & 86.00   &  \textbf{25.23} &  \textbf{68.37} &  \textbf{47.63} &  \textbf{48.17} & 68.87  &  \textbf{66.82} \\
%   \bottomrule

  \toprule
  & \multicolumn{12}{c}{Imbalance Ratio $\tau$ = 0.05} \\
  % & \textbf{Source} & \multicolumn{11}{c}{\textbf{Target}} \\ 
  % \cmidrule(lr){2-2} \cmidrule(lr){3-13}
  & 
  \multicolumn{1}{c}{IN. $\rightarrow$} &
    \multicolumn{1}{c}{Cal.} &
    \multicolumn{1}{c}{OP.} &
    \multicolumn{1}{c}{SC.} &
    \multicolumn{1}{c}{Flw.} &
    \multicolumn{1}{c}{Food.} &
    \multicolumn{1}{c}{FA.} &
    \multicolumn{1}{c}{SUN.} &
    \multicolumn{1}{c}{DTD} &
    \multicolumn{1}{c}{ES.} &
    \multicolumn{1}{c}{UCF.} &
    \multicolumn{1}{c}{Avg}  
  \\
  \midrule
    % {CoOp} & & & & & & & & & & & &                \\
  CoCoOp &  69.93 & 93.18 & 89.88 & 65.09 & 71.02  & 85.21  & 20.02  & 66.89  &44.25  & 44.34& 66.53  &64.64  \\
  % MaPLe &69.53& 93.20 &90.17  & 64.60 & 70.87 & 85.97  & 22.76 &66.37& 46.40  &  47.77 & 66.10  &  \\
  MaPLe & 69.20 &92.23  & 90.34 & 65.17 &  71.54 &  85.03 &  21.39 & 65.57 & 45.84 & 47.83 & 65.34 & 65.03    \\
  % \midrule
% \rowcolor{tabhighlight} Ours & 69.62 & 93.90 &89.70  & 65.43 & 71.90 &  85.77 &  24.02 & 66.57&46.97 &45.10 & 67.00  &  \\
\rowcolor{tabhighlight}
{CoCoOp + NPT}  &  \textbf{70.14} &  \textbf{93.52} &  \textbf{91.56} &  \textbf{66.43} & 71.43 & 85.98 & 22.34 &  \textbf{67.42} & 45.33 & 44.90 &  \textbf{67.24} & 65.62           \\ 
\rowcolor{tabhighlight} {MaPLe + NPT}  & 69.40 & 93.31 & 90.64 & 65.95  &  \textbf{72.62}  &  \textbf{86.19} &  \textbf{22.78} & 66.89 &  \textbf{46.05} &  \textbf{50.25} & 67.12&  \textbf{66.18}    \\
  \bottomrule

  \toprule
  & \multicolumn{12}{c}{Imbalance Ratio $\tau$ = 0.01} \\
  % & \textbf{Source} & \multicolumn{11}{c}{\textbf{Target}} \\ 
  % \cmidrule(lr){2-2} \cmidrule(lr){3-13}
  & 
  \multicolumn{1}{c}{IN. $\rightarrow$} &
    \multicolumn{1}{c}{Cal.} &
    \multicolumn{1}{c}{OP.} &
    \multicolumn{1}{c}{SC.} &
    \multicolumn{1}{c}{Flw.} &
    \multicolumn{1}{c}{Food.} &
    \multicolumn{1}{c}{FA.} &
    \multicolumn{1}{c}{SUN.} &
    \multicolumn{1}{c}{DTD} &
    \multicolumn{1}{c}{ES.} &
    \multicolumn{1}{c}{UCF.} &
    \multicolumn{1}{c}{Avg}  
  \\
  \midrule
  % CoOp&65.70&93.89&91.67&66.93&70.27&89.57&10.93&71.06&48.63&50.07&70.23&63.72 \\
  CoCoOp & 68.42 & 92.45 & 89.21  & 64.34  & 70.32  & 84.13  & 19.88 & 65.07 & 44.09 & 44.18 & 64.31  & 63.80 \\
  % MaPLe  & 67.32 & 91.90 & 90.50 & 61.93 & 67.03  & 85.77  &  21.10 & 61.80  & 45.73 & 47.33 & 66.80    &63.99  \\
  MaPLe & 68.32 &92.63  & 90.07 & 64.38 &  70.37 &  84.77 &  20.47 & 64.53 & 45.17 & 47.03 & 64.40&64.38    \\

  % \midrule
  \rowcolor{tabhighlight}
{CoCoOp + NPT}  & 68.88 & 93.03 &  \textbf{90.12} & 65.22 &70.88 & 85.65 & 20.87 &  \textbf{66.23} &  \textbf{45.34} & 44.89 & 65.91 & 64.81           \\ 
% \rowcolor{tabhighlight} {MaPLe + NPT}  &68.70 & 92.50 & 89.93  & 62.70 & 67.00 &  85.90 &  23.37 & 64.90 & 46.97 & 49.43 & 66.20    & 64.89 \\
\rowcolor{tabhighlight} {MaPLe + NPT}  &  \textbf{68.90} &  \textbf{93.60} & 89.93 &  \textbf{65.28}  &  \textbf{71.77}  &  \textbf{86.21} &  \textbf{21.51} & 66.07 &  44.97 &  \textbf{49.67} &  \textbf{66.20} &  \textbf{65.52}    \\
 
\bottomrule
  
  \toprule
  \end{tabular}}
\end{table*}

  \begin{table}[!t]
  \scriptsize
    \caption{\textbf{Ablation study of each loss module in NPT on three generalization tasks with $\tau=0.01$.}}
    % \vspace{-0.4cm}
    \label{tab:ablation}
    \centering
    % \resizebox{1\linewidth}{!}
    % {
    \renewcommand{\arraystretch}{0.8}

    \begin{tabular}{l|ccc}
    
      \bottomrule
  
      \toprule
    % & \textbf{Source} & \multicolumn{11}{c}{\textbf{Target}} \\ 
    % \cmidrule(lr){2-2} \cmidrule(lr){3-13}
    & 
    \multicolumn{1}{c}{Base-to-Novel} &
      \multicolumn{1}{c}{Cross-Dataset} &
      \multicolumn{1}{c}{Domain Generalization} 
      % &
      % \multicolumn{1}{c}{SC.} &
      % \multicolumn{1}{c}{Flw.} &
      % \multicolumn{1}{c}{Food.} &
      % \multicolumn{1}{c}{FA.} &
      % \multicolumn{1}{c}{SUN.} &
      % \multicolumn{1}{c}{DTD} &
      % \multicolumn{1}{c}{ES.} &
      % \multicolumn{1}{c}{UCF.} &
      % \multicolumn{1}{c}{Avg}  
    \\
    \midrule
    % CoCoOp & 68.31 & &            \\ 
    % + $\mathcal{L}_\text{LC}$& 69.23 & &             \\ 
    % + $\mathcal{L}_\text{MI}$ & 69.89 & &         \\ 
    % \rowcolor{tabhighlight}
    % + NPT & 70.77 & &             \\ 
    % \midrule
    MaPLe & 69.38 & 63.99 & 59.12           \\ 
    + $\mathcal{L}_\text{LC}$& 70.02 & 64.01 & 59.45              \\ 
    + $\mathcal{L}_\text{MI}$ &70.43& 64.32&  59.48            \\ 
    \rowcolor{tabhighlight}
    + NPT & \textbf{71.85}  & \textbf{64.89} & \textbf{59.53}             \\ 
    \bottomrule
  
  \toprule
    \end{tabular} 
    % }
\end{table}

 We further compare the accuracy gains of MaPLE on both novel and base classes before and after integrating NPT in Fig.~\ref{fig:b2n}. 
  Our method significantly outperforms MaPLe in both base and novel classes across all 11 datasets. 
Interestingly, we observe that NPT offers a more substantial improvement in recognizing novel classes, highlighting its potential in challenging generalization scenarios.  
% We plan to delve deeper into this phenomenon in our subsequent research.
 % Surprisingly, on the EuroSAT dataset with two different imbalance ratios, our method significantly improves the accuracy in novel classes by over 10\%. 

  \subsection{Cross-Dataset Transfer}
  We then evaluate the generalization ability of our method on more challenging cross-dataset tasks. In this setting, we learn prompts on ImageNet of 1000 classes with three different imbalance ratios. The effectiveness of the learned prompts is then tested on the other 10 datasets. The results are reported in Table \ref{tab:xd}. The same performance drop phenomenon can be observed in cross-dataset tasks due to data imbalance, and combining NPT with CoCoOp and MaPLe can improve this drop. MaPLe+NPT achieves the best results at all imbalance ratios, demonstrating the great transfer ability of our method.

  \subsection{Domain Generalization}
  % The domain generalization setting evaluates the generalization ability of the model on the target domain that is similar to but different from the source domain. 
  % Zero-shot CLIP introduces no additional training parameters and exhibits great robustness to naturally distribution shifts. Other methods use few samples to train learnable parameters, there is a risk of overfitting the source distribution.   
  We perform experiments with ImageNet as the source domain and evaluate its ability to generalize to four distinct ImageNet variants, which serve as unseen target domains.
  Table \ref{tab:dg} presents the results, indicating that our method significantly improves the performance on the four ImageNet variants with two imbalance ratios. This confirms that our approach enhances the discriminability of the source dataset while preserving its generalization to the target domain.

\subsection{Improvement of NPT on Balanced Datasets}
In Table \ref{tab:b2n}, we showcase the improvement of NPT on balanced datasets under the base-to-novel task. Our NPT can improve generalization for most of the datasets (9/11), and the maximal performance gain is up to 1.3\%. 
Due to space constraints, other experimental results with balanced data ($\tau = 1$) regarding cross-dataset transfer and domain generalization are provided in the Appendix.

\begin{figure}[!t]

  \centering
  \subfloat{
  \includegraphics[width=3in]{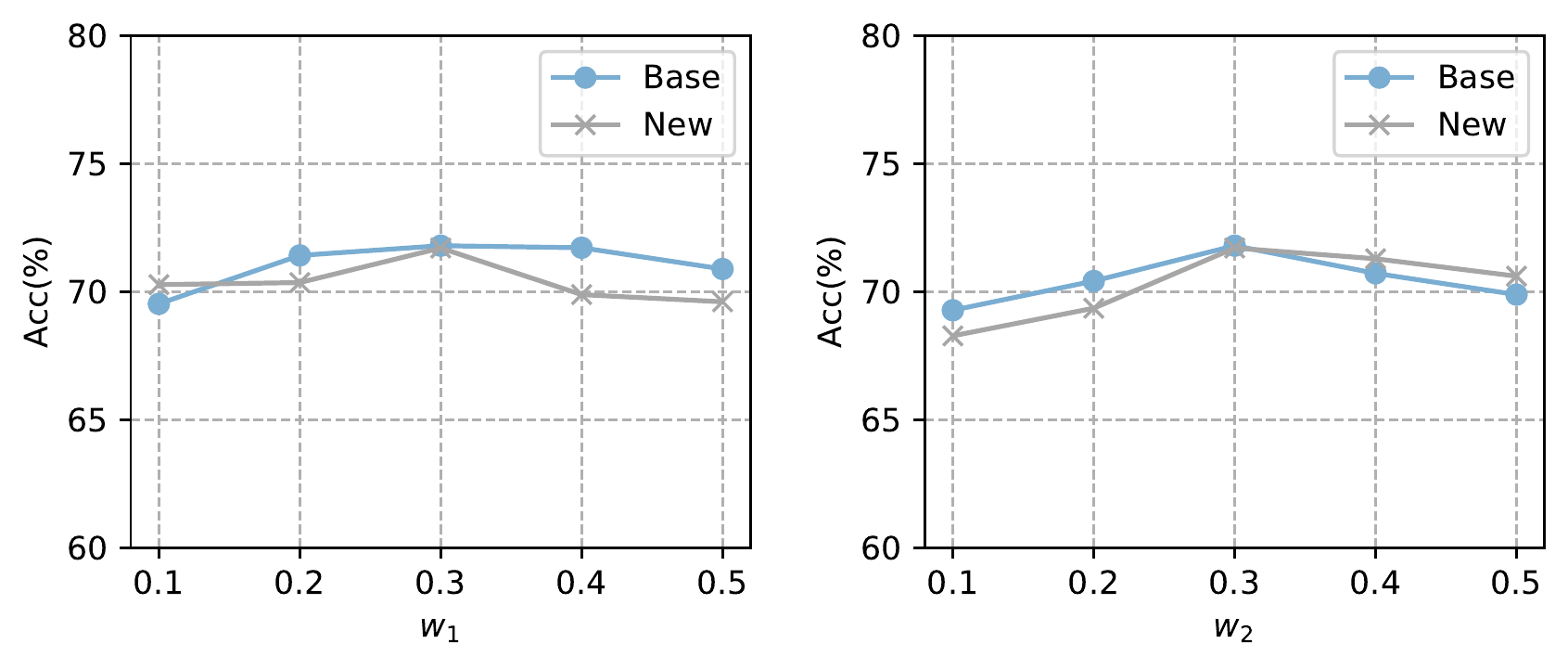}
  }
  % \vspace{-0.4cm}
  \caption{\textbf{Sensitivity analysis of $w_1$ and $w_2$ under base-to-novel task with $\tau = 0.01$.}
  }
  \label{fig:sensitive}
\end{figure}
\subsection{Ablation Study}
\noindent \textbf{Effectiveness of each module.} We conduct ablation experiments on three tasks to assess the effectiveness of the $\mathcal{L}_\text{LCD}$ and $\mathcal{L}_\text{MID}$ of NPT, as presented in Table \ref{tab:ablation}. The results demonstrate that each module has a significant positive impact on the model's performance. Specifically, for the base-to-novel task, the $\mathcal{L}_\text{LCD}$ and $\mathcal{L}_\text{MID}$ modules improve the results by 0.64\% and 1.05\%, respectively. Furthermore, the combination of these modules leads to a remarkable improvement of 2.47\%. These findings suggest that both modules play a crucial role in enhancing the model's generalization ability.

\noindent \textbf{Sensitivity Analysis of $w_1$ and $w_2$.}
In Fig. \ref{fig:framework}, we analyze the impact of the hyperparameters $w_1$ and $w_2$ of Eq. \ref{eq:overal_loss}. The results show that our method's performance remains relatively stable across a broad range of $w_1$ and $w_2$ values, indicating the robustness of our approach to these hyperparameters. This suggests that our method can be effectively applied to a variety of downstream tasks with different parameter settings.

\begin{table}[!t]
  \scriptsize
  \caption{ \textbf{Comparison of NPT with existing approaches on domain generalization task.}}
  % \vspace{-0.4cm}
  \label{tab:dg}
  \centering
  \resizebox{\linewidth}{!}
  {
  \renewcommand{\arraystretch}{0.9}
  \begin{tabular}{l|ccccccc}
  \bottomrule

  % \toprule
  % & \multicolumn{6}{c}{Imbalance Ratio $\tau$ = 1} \\
  % % & \textbf{Source} & \multicolumn{11}{c}{\textbf{Target}} \\ 
  % % \cmidrule(lr){2-2} \cmidrule(lr){3-13}
  % & 
  % \multicolumn{1}{c}{IN. $\rightarrow$} &
  %   \multicolumn{1}{c}{IN.-V2} &
  %   \multicolumn{1}{c}{IN.-S} &
  %   \multicolumn{1}{c}{IN.-A} &
  %   \multicolumn{1}{c}{IN.-R} &
  %   \multicolumn{1}{c}{Avg}  
  % \\
  % \midrule
  % % CLIP & 66.73 & 60.83 & 46.15 & 47.77 & 73.96  & \\
  % CoOp & 71.51 & 64.20 & 47.99 & 49.71 &  75.21 & 59.28\\
  % CoCoOp & 71.02 & 64.07 & 48.75 & 50.63 & 76.18  & 59.91\\
  % % {ProGrad}   &  &  &  &  &   & \\
  % % {KgCoOp}    &  &  &  &  &   & \\
  % MaPLe &  \textbf{70.72} & 64.07 &  \textbf{49.15} & 49.90 & 76.98  &60.23 \\

  % \midrule
  % \rowcolor{tabhighlight} 
  % % CoCoOp+NPT &  &  &  &  &   & \\
  % \rowcolor{tabhighlight} 
  % MaPLe+NPT  & 70.70 &  \textbf{65.03} & 48.27 &  \textbf{50.70}  &  \textbf{77.33}  &  \textbf{60.33}    \\
  % \bottomrule

  \toprule
  & \multicolumn{6}{c}{Imbalance Ratio $\tau$ = 0.05} \\
  % & \textbf{Source} & \multicolumn{11}{c}{\textbf{Target}} \\ 
  % \cmidrule(lr){2-2} \cmidrule(lr){3-13}
  & 
  \multicolumn{1}{c}{IN. $\rightarrow$} &
    \multicolumn{1}{c}{IN.-V2} &
    \multicolumn{1}{c}{IN.-S} &
    \multicolumn{1}{c}{IN.-A} &
    \multicolumn{1}{c}{IN.-R} &
    \multicolumn{1}{c}{Avg}  
  \\
  \midrule
  % CoOp &  &  &  &  &   & \\
  CoCoOp & 68.43 & 62.21 & 46.89  & 48.56 & 75.64  & 60.37 \\
  % {ProGrad}   &  &  &  &  &   & \\
  % {KgCoOp}    &  &  &  &  &   & \\
  MaPLe & 69.20 &62.40  & 48.27 & 50.98 &76.83   &59.62 \\
  \midrule
\rowcolor{tabhighlight} 
CoCoOp+NPT & \textbf{69.65} & \textbf{64.32} & 47.41 & 48.99 & \textbf{77.89}  & \textbf{61.65} \\
\rowcolor{tabhighlight} 
MaPLe+NPT &  {69.40} &  {63.10} &  \textbf{48.83} & \textbf{51.10}  & {77.20}   &  {60.06}    \\
  \bottomrule

  \toprule
  & \multicolumn{6}{c}{Imbalance Ratio $\tau$ = 0.01} \\
  % & \textbf{Source} & \multicolumn{11}{c}{\textbf{Target}} \\ 
  % \cmidrule(lr){2-2} \cmidrule(lr){3-13}
  & 
  \multicolumn{1}{c}{IN. $\rightarrow$} &
    \multicolumn{1}{c}{IN.-V2} &
    \multicolumn{1}{c}{IN.-S} &
    \multicolumn{1}{c}{IN.-A} &
    \multicolumn{1}{c}{IN.-R} &
    \multicolumn{1}{c}{Avg}  
  \\
  \midrule
  % CoOp &  &  &  &  &   & \\
  CoCoOp & 66.21 & 60.35 & 46.13 & 50.25 & 74.56  &57.82 \\
  % {ProGrad}   &  &  &  &  &   & \\
  % {KgCoOp}    &  &  &  &  &   & \\
  MaPLe & 67.32 &  62.33  &47.20 &  \textbf{50.47}  & 76.47 &59.12   \\
  \midrule
\rowcolor{tabhighlight} 
CoCoOp+NPT & 67.57 & 62.04 & 47.55 & 50.38 & 75.88  &58.96 \\
\rowcolor{tabhighlight} 
MaPLe+NPT &  \textbf{68.70} &  \textbf{62.83} &  \textbf{48.53} & 50.07 &   \textbf{76.70}   &  \textbf{59.53}      \\

\bottomrule

\toprule
  \end{tabular}}
\end{table}

\section{Conclusions}
In this paper, we explore the text-to-image representation geometry of CLIP prompt tuning from the perspective of neural collapse. It is found that class imbalance has detrimental effects on representation, thus hurting generalization. We propose Neural-collapse-anchored Prompt Tuning (NPT), a method that optimizes prompts to ensure text and image representations share the same simplex ETF structure. By incorporating language-modality collapse and multi-modality isomorphism regularization terms, NPT enhances V-L models' generalizability. Experiments show that NPT outperforms existing techniques across 11 diverse datasets under base-to-novel, cross-dataset, and domain generalization tasks.

\bibliography{aaai24}

\end{document}